\documentclass[10pt,twocolumn,letterpaper]{article}

\usepackage{cvpr}
\usepackage{times}
\usepackage{graphicx}
\usepackage{amsmath}
\usepackage{amssymb}
\usepackage{subfigure}

\usepackage[pagebackref=true,breaklinks=true,letterpaper=true,colorlinks,bookmarks=false]{hyperref}
\newcommand{\omitme}[1]{}
\usepackage[boxed]{algorithm2e} 

\def\R{\mathbb{R}}

\DeclareMathOperator*{\argmin}{arg\,min}
 \cvprfinalcopy % *** Uncomment this line for the final submission

\def\cvprPaperID{2085} % *** Enter the CVPR Paper ID here

\begin{document}

%%%%%%%%% TITLE
\title{Causal graph-based video segmentation}

\author{Camille Couprie \and Cl\'ement Farabet \and
  \and Yann LeCun \and \\ 
  Courant Institute of Mathematical Sciences\\ 
  New York University, New York, NY 10003, USA  
}

%\cvprfinalcopy

\maketitle
%\thispagestyle{empty}

%%%%%%%%% ABSTRACT
\begin{abstract}
   Numerous approaches in image processing and computer vision are
  making use of super-pixels as a pre-processing step.  Among the
  different methods producing such over-segmentation of an image, the
  graph-based approach of Felzenszwalb and Huttenlocher is broadly
  employed. One of its interesting properties is that the regions are computed in a
  greedy manner in quasi-linear time. The algorithm may be trivially
  extended to video segmentation by considering a video as a 3D
  volume, however, this can not be the case for causal segmentation,
  when subsequent frames are unknown. We propose an efficient video
  segmentation approach that computes temporally consistent pixels in a
  causal manner, filling the need for causal and real time applications.
\end{abstract}

%%%%%%%%% BODY TEXT
\section{Introduction}

A segmentation of video into consistent spatio-temporal segments is a
largely unsolved problem. While there have been attempts at video
segmentation, most methods are non causal and non real-time.  This
paper proposes a fast method for real time video segmentation,
including semantic segmentation as an application.
 
An large number of approaches in computer vision makes use of
super-pixels at some point in the process. For example, semantic
segmentation  \cite{farabet2013pami}, geometric
context indentification \cite{geometricContext2005}, extraction of support
relations between object in scenes \cite{Silberman:ECCV12}, etc. Among
the most popular approach for super-pixel segmentation, two types of
methods are distinguishable. Regular shape super-pixels may be produced
using normalized cuts \cite{Shi97normalizedcuts} for instance. More
object -- or part of object -- shaped super-pixels can be generated from
watershed based approaches. In particular the method of Felzenswalb
and Huttenlocher \cite{DBLP:journals/ijcv/FelzenszwalbH04} produces such results.

It is a real challenge to obtain a decent delineation of objects from
a single image.  When it comes to real-time data analysis, the problem
is even more difficult. However, additional cues can be used to
constrain the solution to be temporally consistent, thus helping to
achieve better results. Since many of the underlying algorithms are in
general super-linear, there is often a need to reduce the
dimensionality of the video.  To this end, developing low level vision
methods for video segmentation is necessary. Currently, most video
processing approaches are non-causal, that is to say, they
make use of future frames to segment a given frame, sometimes requiring
the entire video \cite{CVPR2010hierarchicalvideo}. This prevents their use for real-time applications.

Some approaches have been designed to address the causal video
segmentation problem
\cite{DBLP:conf/eccv/Paris08,Miksik_2012_7294}. \cite{DBLP:conf/eccv/Paris08}
makes use of the mean shift method \cite{Comaniciu02meanshift}. As
this method works in a feature space, it does not necessary cluster
spatially consistent super-pixels. A more recent approach,
specifically applied for semantic segmentation, is the one of Miksik
{\it et al.} \cite{Miksik_2012_7294}. The work of \cite{Miksik_2012_7294} is employing an optical flow method to enforce
the temporal consistency of the semantic segmentation. Our approach is
different because it aims to produce super-pixels, and possibly uses
the produced super-pixels for smoothing semantic segmentation
results. Furthermore, we do not use any optical flow pre-computation that
would prevent us having real time performances on a CPU.

Some works use the idea of enforcing some consistency between
different segmentations
\cite{Glasner2011CVPR,DBLP:conf/icmcs/LeeOH05,DBLP:conf/cvpr/JoulinBP12,meyer2003icip}. \cite{Glasner2011CVPR}
formulates a co-clustering problem as a Quadratic Semi-Assignment
Problem. However solving the problem for a pair of images takes about
a minute.  Alternatively, \cite{DBLP:conf/icmcs/LeeOH05} and \cite{meyer2003icip} identify the
corresponding regions using graph matching techniques. The approach of
\cite{meyer2003icip} is only illustrated on very coarse image segmentation, and the 
graph matching is performed between graphs that contain a dozen of regions. In
both cases, the number of tracked regions is limited in the
experiments to a small amount.

The idea developped in this paper is to perform independent
segmentations and match the produced super-pixels to define markers.
The markers are then used to produce the final segmentation by
minimizing a global criterion defined on the image.  The graph used in
the independent segmentation part is reused in the final segmentation
stage, leading thus to gains in speed, and real-time performances on a
single core CPU.

\section{Method}

Given a segmentation $S_t$ of an image at time $t$, we wish to compute
a segmentation $S_{t+1}$ of the image at time $t+1$ which is
consistent with the segments of the result at time $t$.

\subsection{Independent image segmentation}

\label{sec:independent}
 
 The super-pixels produced by
 \cite{DBLP:journals/ijcv/FelzenszwalbH04} have been shown to satisfy
 the global properties of being not too coarse and not too fine
 according to a particular region comparison function. In order to
 generate superpixels close to the ones produced by
 \cite{DBLP:journals/ijcv/FelzenszwalbH04}, we first generate
 independent segmentations of the 2D images using
 \cite{DBLP:journals/ijcv/FelzenszwalbH04}.  We name these
 segmentations $S'_{1}, ..., S'_{t}$. The principle of segmentation is
 fairly simple. We define a graph $G_t$, where the nodes correspond to
 the image pixels, and the edges link neighboring nodes in
 8-connectivity. The edge weights $\omega_{ij}$ between nodes $i$ and
 $j$ are given by a color gradient of the image.

%\begin{equation}
%\omega_{ij} = \sqrt{(r_i-r_j)^2+(g_i-g_j)^2+(b_i-bj)^2}
%\label{eq:pairwise_weights}
%\end{equation}

 A Minimum Spanning Tree (MST) is
 build on $G_t$, and a region merging criteria is defined.
Specifically, two regions $X$ and $Y$ are merged when:
{\small
\begin{equation}
Diff(X,Y) \leq \min \left( Int(X) + \frac{k}{|X|} , Int(Y) + \frac{k}{|Y|} \right),
\end{equation}
}
where $k$ is a parameter allowing to prevent the merging of large
regions. The internal difference $Int(X)$ of a region $X$ is the
highest weight of an edge linking two vertices of $X$ in the MST. The
difference $Diff(X,Y)$ between two neighboring regions $X$ and $Y$ is
the smallest weight of an edge that links $X$ to $Y$.\\

Once an image is independently segmented, resulting in $S'_{t+1}$, we then face the question of the propagation of the temporal
consistency given the non overlapping contours of $S_t$ and $S'_{t+1}$.

Our solution is the development of a cheap graph matching technique to
obtain correspondences between segments from $S_t$ and these of
$S'_{t+1}$. This first step is described in Section
\ref{graphmatch}. We then mine these correspondences to create markers
(also called seeds) to compute the final labeling $S_{t+1}$ by solving
a global optimization problem. This second step is detailed in Section
\ref{sec:optim}.

\subsection{Graph matching procedure} 

\label{graphmatch}
\begin{figure}[htb]
\begin{center}
\includegraphics[width=0.5\textwidth]{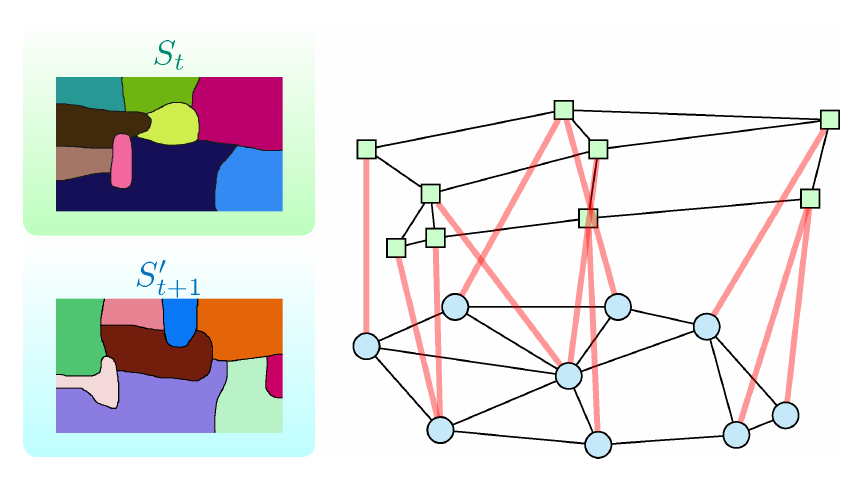}
\end{center}
\caption{Illustration of the graph matching procedure}
\label{fig:problem}
\end{figure}

The basic idea is to use the segmentation $S_t$ and segmentation
$S'_{t+1}$ to produce markers before a final segmentation of image
at time $t+1$. Therefore, in the process of computing a new segmentation $S_{t+1}$, a graph $G$ is defined.
The vertices of $G$ comprises to two sets of vertices: $V_t$
that corresponds to the set of regions of $S_{t}$ and $V'_{t+1}$ that
corresponds to the set of regions of $S'_{t+1}$. Edges link regions
characterised by a small distance between their centroids. The edges weights between vertex $i\in V_t$ and $j\in V'_{t+1}$ are given
by a similarity measure taking into account distance and differences between shape and appearance

\begin{equation}
w_{ij}= \frac {(|r_i|+|r_j|) d(c_i,c_j) }{|r_i\cap r_j|} + a_{ij},
\end{equation}

where $|r_i|$ denotes the number of pixels of region $r_i$, $|r_i\cap~r_j|$ the number of pixels present in $r_i$ and $r_j$ with aligned
centroids, and $a_{ij}$ the appearance difference of regions $r_i$ and
$r_j$. In our experiments $a_{ij}$ was defined as the difference
between mean color intensities of the regions.

The graph matching procedure is illustrated in Figure
\ref{fig:problem} and produces the following result: For each region
of $S'_{t+1}$, its best corresponding region in image $S_t$ is
identified. More specifically, each node $i$ of $V_t$ is associated
with the node $j$ of $V'_{t+1}$ which minimizes
$w_{ij}$. Symmetrically, for each region of $S_{t}$, its best
corresponding region in image $S'_{t+1}$ is identified, that is to say
each node $i$ of $V'_{t+1}$ is associated with the node $j$ of $V_{t}$
which minimizes $w_{ij}$.

\subsection{Final segmentation procedure}

\label{sec:optim}

The final segmentation $S_{t+1}$ is computed using a minimum spanning
forest procedure. This seeded segmentation algorithm that produces
watershed cuts \cite{CoustyBNC09} is strongly linked to global energy
optimization methods such as graph-cuts \cite{Allene,Couprie2011PAMI}
as detailed in Section \ref{sec:guaranties}. In addition to
theoretical guaranties of optimality, this choice of algorithm is
motivated by the opportunity to reuse the sorting of edges that is
performed in \ref{sec:independent} and constitutes the main
computational effort. Consequently, we reuse here the graph
$G_{t+1}(V,E)$ built for the production of independent segmentation
$S'_{t+1}$.

\restylealgo{algoruled}
\begin{algorithm}{}
\label{alg:MSF}
\dontprintsemicolon \SetVline 

\KwData{A weighted graph $G(V, E)$ and a set of labeled nodes makers $L$. 
Nodes of $V \setminus L$ have unknown labels initially.}

\KwResult{A labeling $x$
associating a label to each vertex.}

Sort the edges of $E$ by increasing order of weight. 

\While{any node has an unknown label}{Find the edge 
$e_{ij}$ in $E$ of minimal weight; 

\If{$v_i$ or $v_j$ have unknown label values}
{Merge $v_i$ and $v_j$ into a single node, such that
  when the value for this merged node becomes known, all merged
  nodes are assigned the same value of $x$ and considered known.}
}
\caption{Minimum Spanning Forest algorithm}
\end{algorithm}

The minimum spanning forest algorithm is recalled in Algorithm \ref{alg:MSF}.
The seeds, or markers, are defined using the regions correspondences computed in the previous section,
according to the procedure detailed below. 
For each segment $s'$ of $S'_{t+1}$ four cases may appear: 
\begin{enumerate}
\item  $s'$ has one and only one matching region $s$ in $S_t$: propagate
  the label $l_s$ of region $s$. All nodes of $s'$ are labeled with the label $l_s$ of region $s$.
\item $s'$ has several corresponding regions $s_1, ..., s_r$:
  propagate seeds from $S_t$. The coordinates of regions $s_1, ...,
  s_r$ are centered on region $s'$. The labels of regions $s_1, ...,
  s_r$ whose coordinates are in the range of $s'$ are propagated to
  the nodes of $s'$.
\item $s'$ has no matching region : The region is labeled by the label $l_s'$ itself.
\item If none of the previous cases is fulfilled, it means that $s'$
  is part of a larger region $s$ in $S_t$. If the size of $s'$ is
  small, a new label is created. Otherwise, the label $l_s$ is
  propagated in $s'$ as in case 1.
\end{enumerate}

Before applying the minimum spanning forest algorithm, a safety test
is performed to check that the map of produced markers does not differ
two much from the segmentation $S'_{t+1}$. If the test shows large
differences, an eroded map of $S'_{t+1}$ is used to correct
the markers.

\begin{figure}[htb]
\begin{center}
\begin{tabular}{cc}
\includegraphics[width=0.22\textwidth]{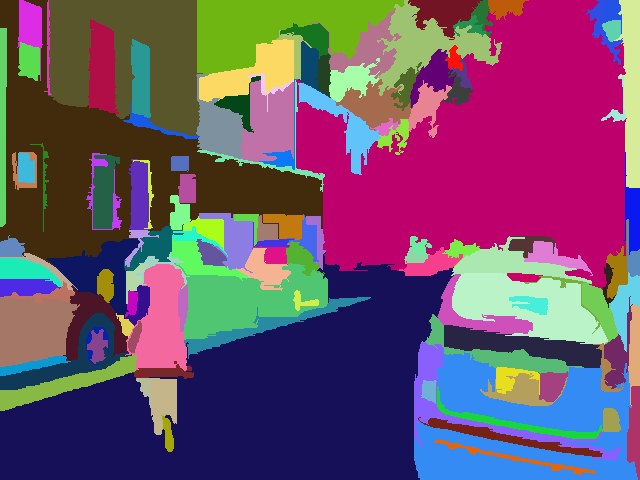}&\includegraphics[width=0.22\textwidth]{figure2a.png}\\
\includegraphics[width=0.22\textwidth]{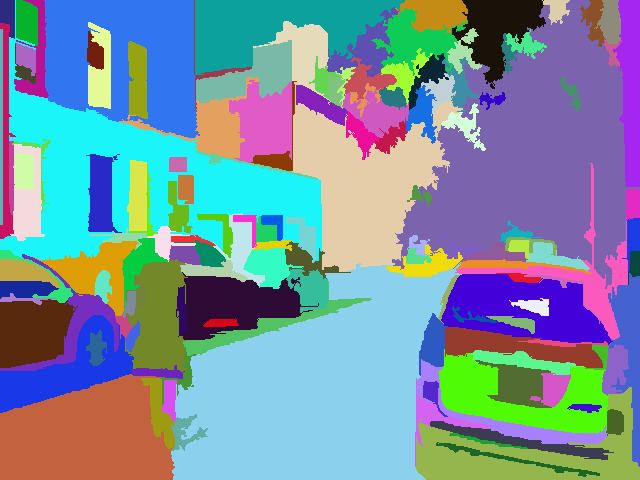}&\includegraphics[width=0.22\textwidth]{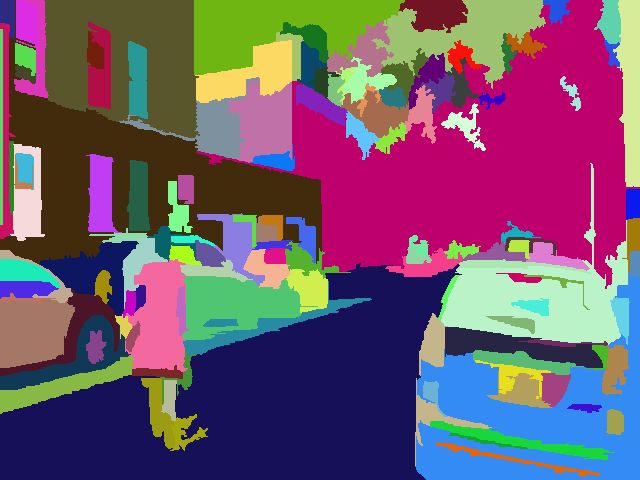}\\
\includegraphics[width=0.22\textwidth]{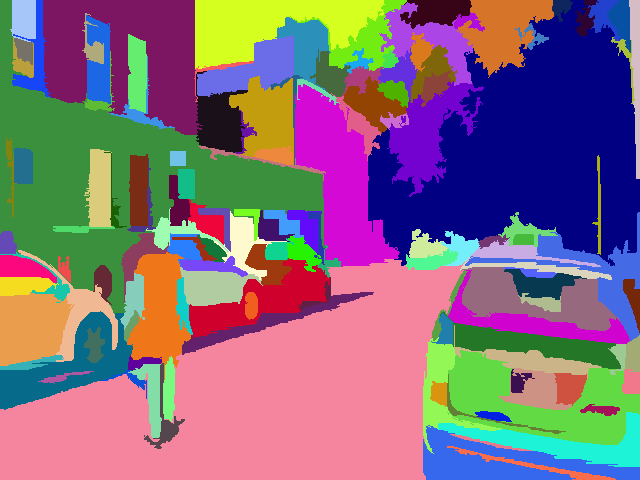}&\includegraphics[width=0.22\textwidth]{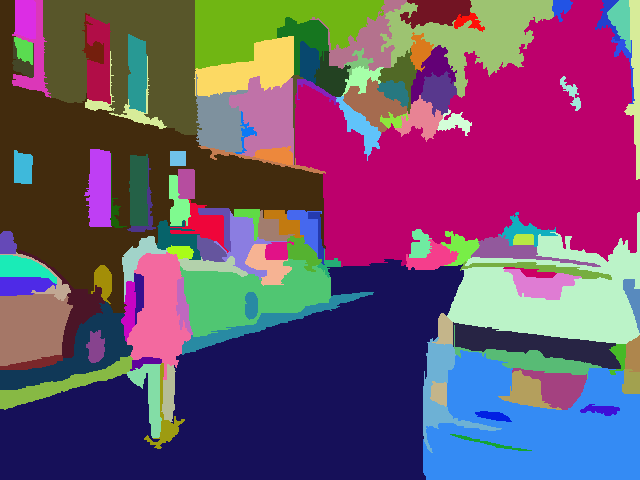}\\
Independent segmentations  & Temporally consistent segm-\\
$S'_1, S'_2$ and $S'_3$  & entations $S_1 (=S'_1), S_2$, and $S_3$
\end{tabular}
\end{center}
\caption{Segmentation results on 3 consecutive frames of the NYU-Scene dataset.}
\label{fig:seg-result}
\end{figure}

\subsection{Global optimization guaranties}

\label{sec:guaranties}
  
Several graph-based segmentation problems, including minimum spanning
forests, graph cuts, random walks and shortest paths have recently
been shown to belong to a common energy minimization framework \cite{Couprie2011PAMI}. 
The considered problem is to find a labeling $x^*\in \R^{|V|}$ defined on the nodes of a graph that minimizes 

\begin{align}
\begin{gathered}
E(x) = \sum_{ e_{ij} \in E} w_{ij}^p|x_j-x_i|^q + \sum_{v_i \in V} w_{i}^p |l_i-x_i|^q,
\label{eqpower}
\end{gathered}
\end{align}

where $l$ represents a given configuration and $x$ represents the
target configuration. The result of $\lim_{p\rightarrow\infty}
\argmin_x E(x)$ for values of $q\geq1$ always produces a cut by
maximum (equivalently minimum) spanning forest. The reciprocal is also true if the weights of the graph are all different. 

In the case of our application, the pairwise weights $w_{ij}$ is given
by an inverse function of the original weights $\omega_{ij}$. The
pairwise term thus penalizes any unwanted high-frequency content in
$x$ and essentially forces $x$ to vary smoothly within an object,
while allowing large changes across the object boundaries. The second
term enforces fidelity of $x$ to a specified configuration $l$,
$w_{i}$ being the unary weights enforcing that fidelity.

The enforcement of markers $l_s$ as hard constrained may be viewed as follows: 
A node of label $l_s$ is added to the graph, and linked
to all nodes $i$ of $V$ that are supposed to be marked. The unary
weights $\omega_{i,l_s}$ are set to arbitrary large values in order
to impose the markers.

\begin{figure*}[htb]
\begin{center}
\begin{tabular}{ccc}
\includegraphics[width=0.31\textwidth]{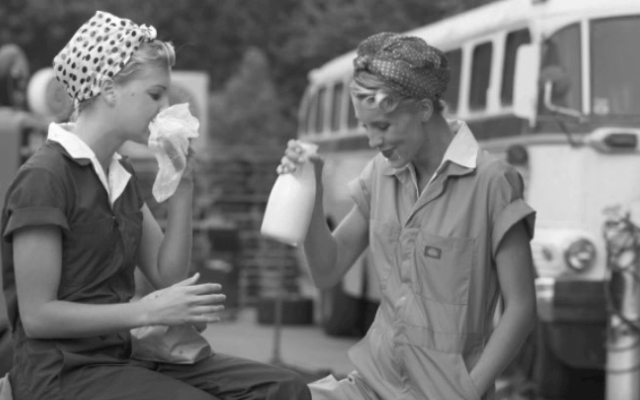}&
\includegraphics[width=0.31\textwidth]{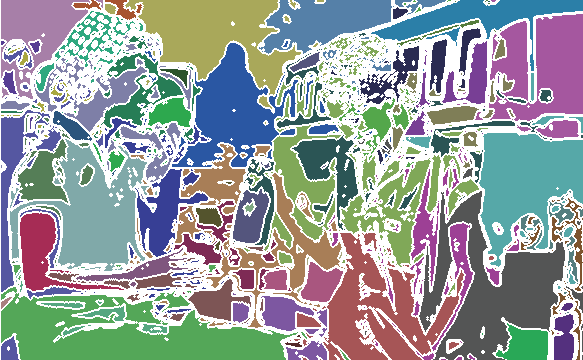}&
\includegraphics[width=0.31\textwidth]{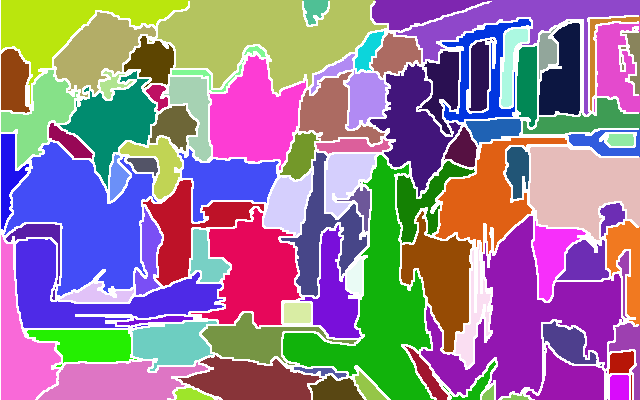}\\
\includegraphics[width=0.31\textwidth]{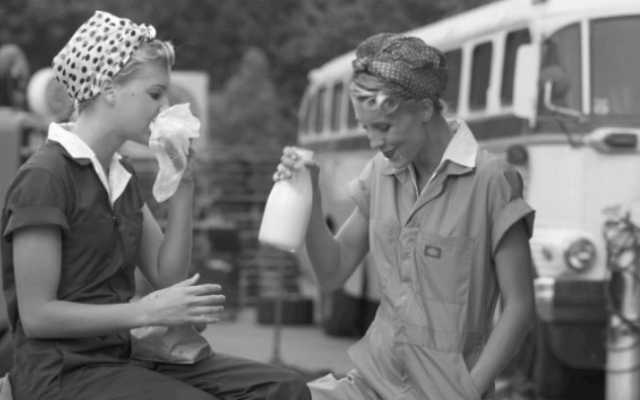}&
\includegraphics[width=0.31\textwidth]{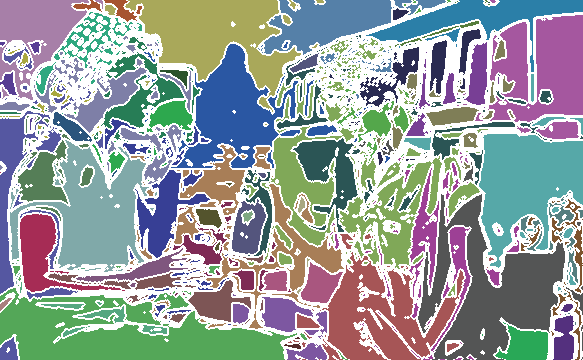}&
\includegraphics[width=0.31\textwidth]{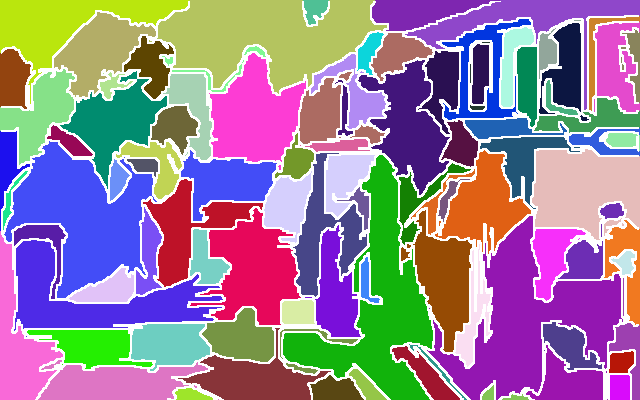}\\
\end{tabular}
\end{center}
\caption{Comparison with the mean-shift segmentation method of Paris \cite{DBLP:conf/eccv/Paris08} on Frame 19 and 20. $k= 200, \delta=400, \sigma = 0.5$.}
\label{fig:mean-shift}
\end{figure*}

\subsection{Applications to optical flow and semantic segmentation}

\label{sec:opt}

An optical flow map may be easily estimated from two successive
segmentations $S_{t}$ and $S_{t+1}$.  For each region $r$ of
$S_{t+1}$, if the label of $r$ comes from a label present in a region
$s$ of the segmentation $S_{t}$, the optical flow in $r$ is computed as the
distance between the centroid of $r$ and the centroid of $s$.  The
optical flow map may be used as a sanity check for region tracking
applications. By principle, a video sequence will not contain
displacements of objects greater than a certain value.

For each superpixel $s$ of $S_{t+1}$, if the label of region $s$ comes
from the previous segmentation $S_t$, then the semantic prediction from
$S_t$ is propagated to $S_{t+1}$. Otherwise, in case the label of $s$
is a new label, the semantic prediction is computed using the
prediction at time $t+1$. As some errors may appear in the regions tracking, the labels of regions 
 that have inconsistent large values in optical flow maps are not propagated.     
  
For the specific task of semantic segmentation, results can be improved by exploiting the contours of the recognized objects. Semantic contours such as for example transition between a building
and a tree for instance, might not be present in the gradient of the
raw image. Thus, in addition to the pairwise weights $\omega$
described in Section \ref{sec:independent}, \omitme{given in equation
  \eqref{eq:pairwise_weights}} we add a constant in the presence of a
semantic contour.

%%%%%%%%%%%%%%%%%%%%%%%%%%%%%%%%%%%%%%%%%%%%%%%%%%%%%%%%%%%%%%%%%%%%
\section{Results} 

We now demonstrate the efficiency and versatility of our approach by
applying it to different problems: simple super-pixel segmentation,
semantic scene labeling, and optical flow.

Following the implementation of
\cite{DBLP:journals/ijcv/FelzenszwalbH04}, we pre-process the images
using a Gaussian filtering step with a kernel of variance $\sigma$ is
employed. A post-processing step that removes regions of small size,
that is to say below a threshold $\delta$ is also performed. As in
\cite{DBLP:journals/ijcv/FelzenszwalbH04}, we denote the scale of
observation parameter by $k$.

\subsection{Super-pixel segmentation}

Experiments are performed on two different types of videos: videos
where the camera is static, and videos where the camera is moving.
The robustness of our approach to large variations in the region sizes
and large movements of camera is illustrated on Figure~\ref{fig:seg-result}.

A comparison with the temporal mean shift segmentation of Paris
\cite{DBLP:conf/eccv/Paris08} is displayed at Figure
\ref{fig:mean-shift}. The super-pixels produced by the
\cite{DBLP:conf/eccv/Paris08} are not spatially consistent as the
segmentation is performed in the feature (color) space in their
case. Our approach is slower, although qualified for real-time
application but computes only spatially consistent super-pixels.

\omitme{ Another application of this work can be biological imagery,
  as illustrated in Figure \ref{fig:cell-division}.  The segmentation
  of a video sequence of cell division is accurately performed, and
  specific times of cell division can be detected using our fully
  automatic procedure.

\begin{figure}[htb]
\begin{center}
\scriptsize{
\begin{tabular}{cccc}
\includegraphics[width=0.1\textwidth]{images/Frame00091.png}&
\includegraphics[width=0.1\textwidth]{images/Frame00099.png}&
\includegraphics[width=0.1\textwidth]{images/Frame00100.png}&
\includegraphics[width=0.1\textwidth]{images/Frame00108.png}\\
\includegraphics[width=0.1\textwidth]{images/out00091.png}&
\includegraphics[width=0.1\textwidth]{images/out00099.png}&
\includegraphics[width=0.1\textwidth]{images/out00100.png}&
\includegraphics[width=0.1\textwidth]{images/out00108.png}\\
Frame 91 & Frame 99 & Frame 100 & Frame 108
\end{tabular}
}
\end{center}
\caption{ Cell division. $k = 250, \delta = 1000, \sigma = 0.6$ }
\label{fig:cell-division}
\end{figure}}

\begin{figure*}[htb]
\begin{center}
\begin{tabular}{ccccc}
\includegraphics[width=0.18\textwidth]{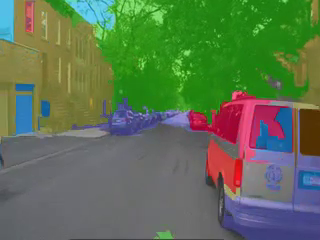}&
\includegraphics[width=0.18\textwidth]{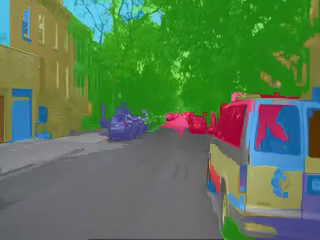}&
\includegraphics[width=0.18\textwidth]{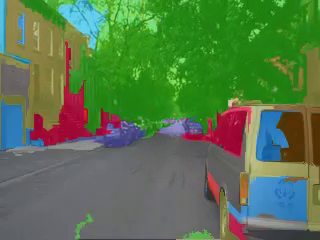}&
\includegraphics[width=0.18\textwidth]{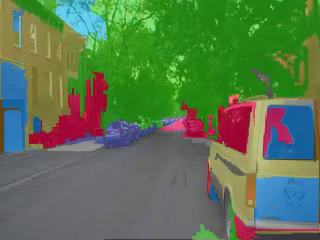}&
\includegraphics[width=0.18\textwidth]{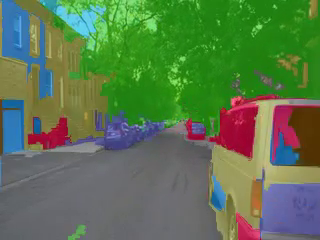}
\end{tabular}\\
(a) Independent segmentations with no temporal smoothing \\
\begin{tabular}{ccccc}
\includegraphics[width=0.18\textwidth]{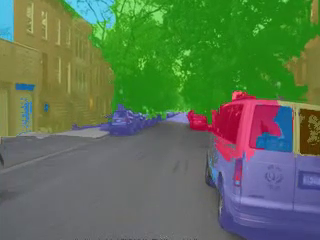}&
\includegraphics[width=0.18\textwidth]{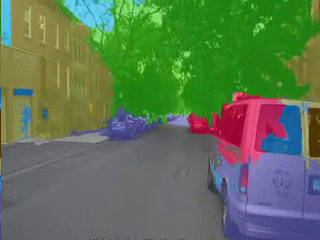}&
\includegraphics[width=0.18\textwidth]{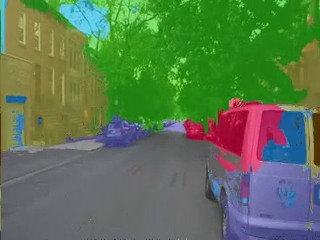}&
\includegraphics[width=0.18\textwidth]{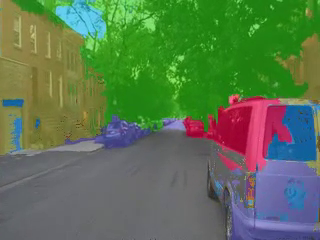}&
\includegraphics[width=0.18\textwidth]{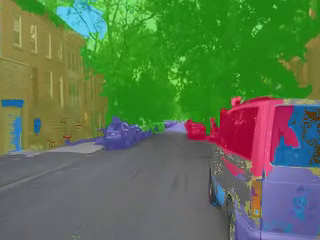}
\end{tabular}\\
(b) Result using the temporal smoothing method of \cite{Miksik_2012_7294}\\
\begin{tabular}{ccccc}
\includegraphics[width=0.18\textwidth]{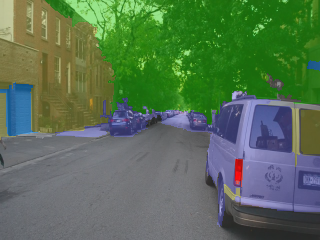}&
\includegraphics[width=0.18\textwidth]{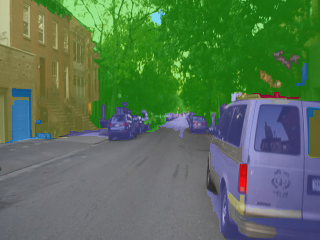}&
\includegraphics[width=0.18\textwidth]{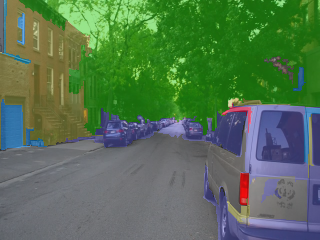}&
\includegraphics[width=0.18\textwidth]{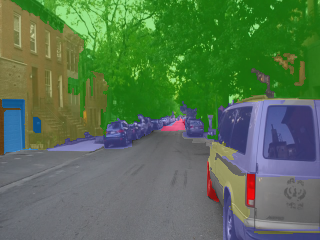}&
\includegraphics[width=0.18\textwidth]{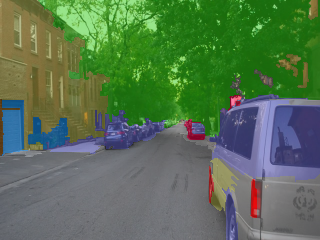}\\
\end{tabular}\\
(c) Our temporally consistent segmentation \\
\bigskip 
\begin{tabular}{ccccccc}
\begin{tabular}{l}  Legend: \\
\includegraphics[width=0.025\textwidth]{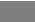} awning 
\end{tabular}&
\begin{tabular}{l}
\includegraphics[width=0.025\textwidth]{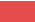} balcony \\
\includegraphics[width=0.025\textwidth]{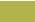} building 
\end{tabular}&
\begin{tabular}{l}
\includegraphics[width=0.025\textwidth]{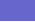} car \\
\includegraphics[width=0.025\textwidth]{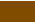} door
\end{tabular}&
\begin{tabular}{l}
\includegraphics[width=0.025\textwidth]{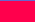} person \\
\includegraphics[width=0.025\textwidth]{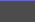} road 
\end{tabular}&
\begin{tabular}{l}
\includegraphics[width=0.025\textwidth]{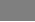} sidewalk\\
\includegraphics[width=0.025\textwidth]{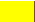} sun 
\end{tabular}&
\begin{tabular}{l}
\includegraphics[width=0.025\textwidth]{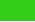} tree \\
\includegraphics[width=0.025\textwidth]{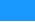} window 
\end{tabular}&
\end{tabular}

\end{center}
\caption{Comparison with the temporal smoothing method of \cite{Miksik_2012_7294}. Parameters used: $k= 1200, \delta=100, \sigma = 1.2$.}
\label{fig:semantic}
\end{figure*}

\subsection{Semantic scene labeling}

We suppose that we are given a noisy semantic labeling for each
frame. In this work we used the semantic predictions of \cite{farabet2013pami}.

We compare our results with the results of \cite{Miksik_2012_7294} on
the NYU-Scene dataset. The dataset consists in a video sequence of 73
frames provided with a dense semantic labeling and ground truths.  The
provided dense labeling being performed with no temporal exploitation,
it suffers from sudden large object appearances and disappearances. As
illustrated in Figure \ref{fig:semantic} our approach reduces this
effect, and improves the classification performance of more than 5\%
as reported in Table \ref{tab}. We also display results on another video in Figure \eqref{fig:flow}.

\begin{table}[htb]
\begin{center}
\begin{tabular}{|c|c|c|c|}
\hline
 & Frame & Miksik & Our \\
& by frame & et al.\cite{Miksik_2012_7294} & method \\
\hline
Accuracy  & 71.11 & 75.31 & 76.27 \\
\hline
\#Frames/sec &       & $1.33^*$  & 10.5 \\
%MPI-VehicleScenes & 93.27 & 93.76 & \\
\hline
\end{tabular}
\vspace{2ex}
\caption{Overall pixel accuracy ($\%$) for the semantic segmentation task on the NYU Scene video. $^*$Note that the reported timing does not take into account the optical flow computation needed by \cite{Miksik_2012_7294}.}
\end{center}
\label{tab}
\end{table}

\subsection{Optical flow}

\begin{figure*}[hbt]
\begin{center}
\subfigure[Frame 9]{\includegraphics[width=0.24\textwidth]{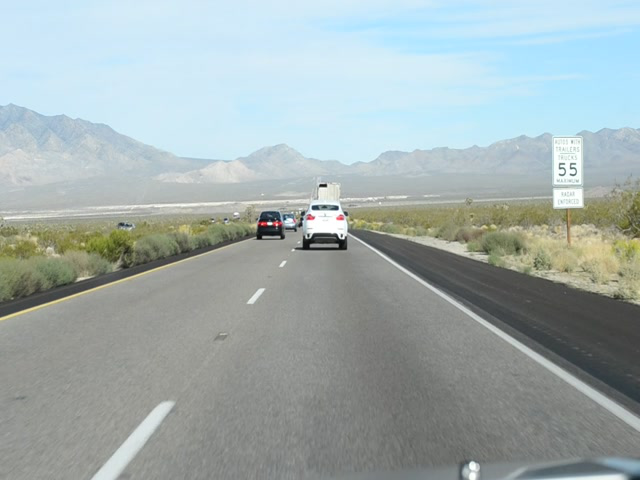}}~
\subfigure[Frame 10]{\includegraphics[width=0.24\textwidth]{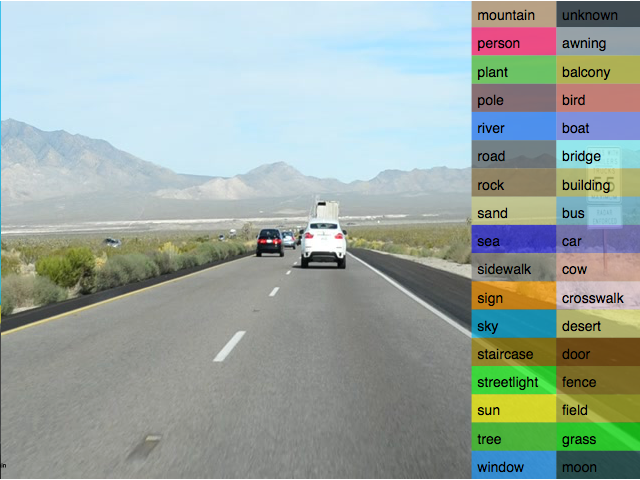}}~
\subfigure[Segmentation $S_9$]{\includegraphics[width=0.24\textwidth]{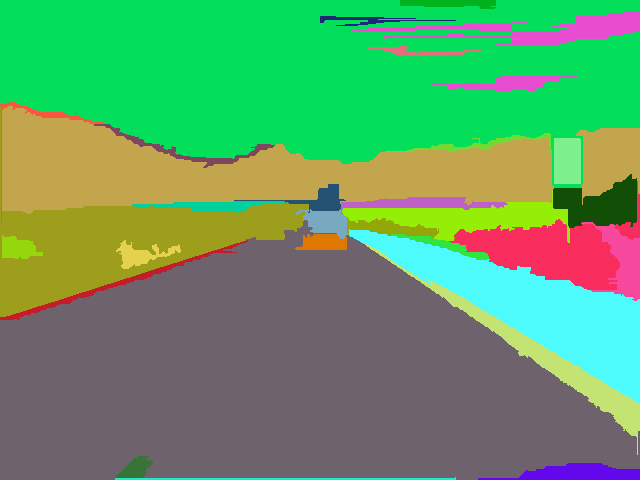}}~
\subfigure[Segmentation $S_{10}$]{\includegraphics[width=0.24\textwidth]{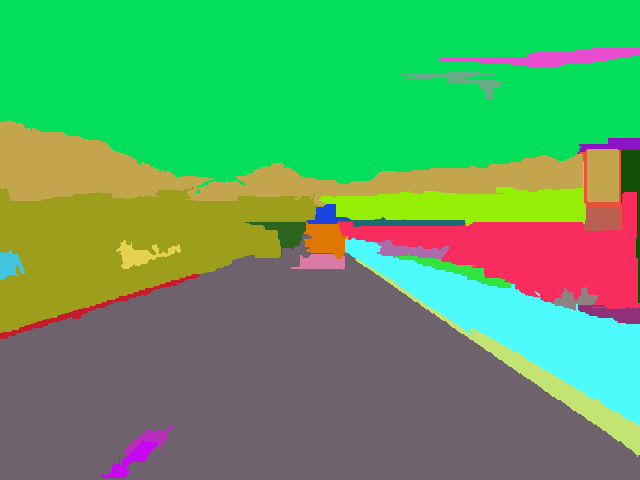}}\\
\subfigure[Optical flow]{\includegraphics[width=0.26\textwidth]{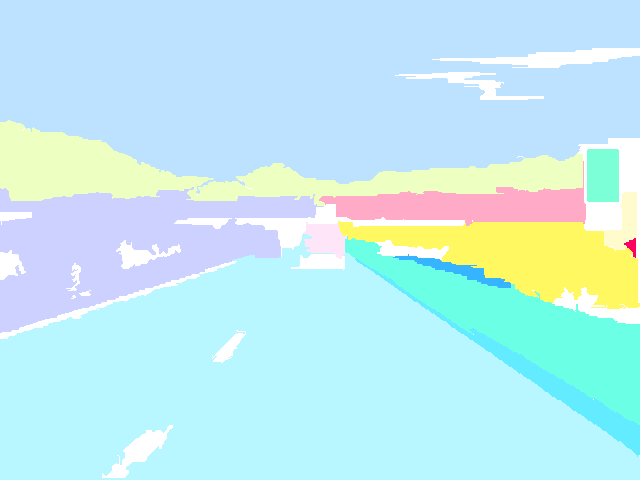}}~
\subfigure[Legend]{\includegraphics[width=0.195\textwidth]{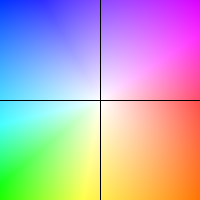}}~
\subfigure[Frame by frame segm.]{\includegraphics[width=0.26\textwidth]{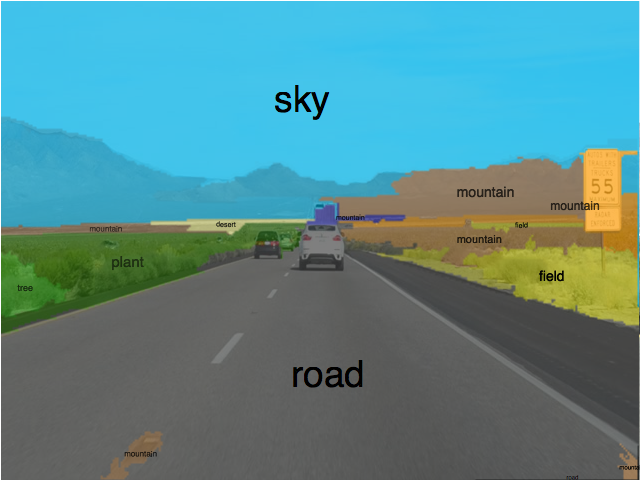}}~
\subfigure[Our semantic segm.]{\includegraphics[width=0.26\textwidth]{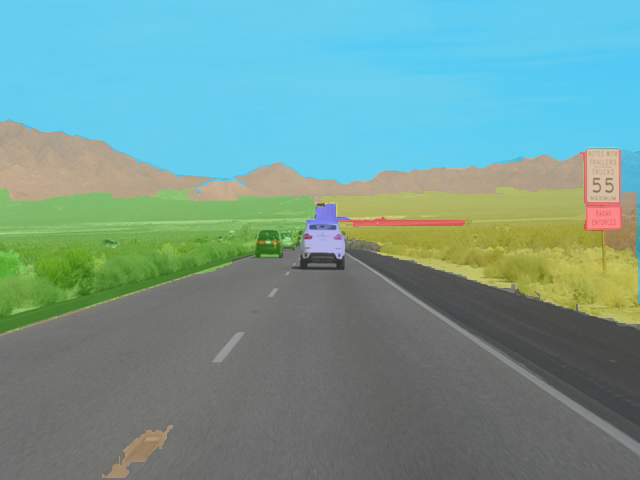}}~
\end{center}
\caption{By matching regions, our method can be used to derive an optical flow (e). Parameters used: $k=200, \delta = 400, \sigma=0.81$  }
\label{fig:flow}
\end{figure*}

As detailed in Section \ref{sec:opt}, we compute the optical flow maps
between subsequent frames of videos.  An example of result is shown in
Figure \ref{fig:flow}, that illustrates the accurate detection of a
large move of the camera from the Frames 9 to 10 of a video taken on a
highway.

\subsection{Computation time}

The experiments were performed on a laptop with 2.3 GHz intel core
i7-3615QM, Memory 8Go DDR3 1600 MHz. Our method is implemented on CPU
only, in C/C++, and makes use of only one core of the processor.
Super-pixel segmentations take 0.1 seconds per image of size
$320\times240$ and 0.4 seconds per image of size $640\times380$, thus
demonstrating the scalability of the pipeline. All computations are
included in the reported timings.

The timings of the temporal smoothing method of Miksik {\it et
  al.}\cite{Miksik_2012_7294} are reported in Table \ref{tab}. We note
that the processor used for the reported timings of
\cite{Miksik_2012_7294} has similar characteristics as
ours. Furthermore, Mistik {\it et al.}  use an optical flow procedure
that takes only 0.02 seconds per frame when implemented on GPU, but
takes seconds on CPU. Our approach is thus more adapted to real time
applications for instance on embedded devices where a GPU is often not
available.

%%%%%%%%%%%%%%%%%%%%%%%%%%%%%%%%%%%%%%%%%%%%%%%%%%%%%%%%%%%%%%%%%
\section{Conclusion}

The proposed approach employs a graph matching technique to produce
markers used in a global optimization procedure for video
segmentation. Unlike many video segmentation techniques, our algorithm
is causal -- which is a required property for real-time applications
-- and does not require any computation of optical flow. Our
experiments on challenging videos show that the obtained super-pixels
are robust to large camera or objects displacement. Their use in
semantic segmentation applications demonstrate that significant gains
can be achieved and lead to state-of-the-art results. Furthermore, by
being 8 times faster than the competing method for temporal smoothing
of semantic segmentation, and up to 25 times faster if the use of GPU
is not available, the proposed approach has by itself a practical
interest.

\section*{Acknowledgments}

The authors would like to thank Laurent Najman for fruitful discussions. 
{\small
\bibliographystyle{ieee}
\bibliography{cvpr2013video_segm}
}

\end{document}